\documentclass[journal, 12pt]{IEEEtran}
\onecolumn
\usepackage{setspace}
\usepackage{amsmath}
\usepackage[margin=1in]{geometry} 
\usepackage{graphicx} 
\usepackage{algorithm}
\usepackage{algpseudocode}
\usepackage[justification=centering]{caption}
\usepackage[numbers]{natbib}
\usepackage{mdframed} 
 \usepackage{url} 
\usepackage{amssymb}  

\begin{document}

\title{Communication-Efficient Federated Learning via Clipped Uniform Quantization}

\author{ Zavareh Bozorgasl,~\IEEEmembership{Member,~IEEE} \thanks{Z. Bozorgasl (zavarehbozorgasl@u.boisestate.edu) and H. Chen (haochen@boisestate.edu) are with the Department
of Electrical and Computer Engineering, Boise State University, Boise,
ID, 83712.}, Hao~Chen,~\IEEEmembership{Member,~IEEE}
\thanks{},
}

\maketitle


\begin{abstract}
This paper presents a novel approach to enhance communication efficiency in federated learning through clipped uniform quantization. By leveraging optimal clipping thresholds and client-specific adaptive quantization schemes, the proposed method significantly reduces bandwidth and memory requirements for model weight transmission between clients and the server while maintaining competitive accuracy. We investigate the effects of symmetric clipping and uniform quantization on model performance, emphasizing the role of stochastic quantization in mitigating artifacts and improving robustness. Extensive simulations demonstrate that the method achieves near-full-precision performance with substantial communication savings. Moreover, the proposed approach facilitates efficient weight averaging based on the inverse of the mean squared quantization errors, effectively balancing the trade-off between communication efficiency and model accuracy. Moreover, in contrast to federated averaging, this design obviates the need to disclose client-specific data volumes to the server, thereby enhancing client privacy. Comparative analysis with conventional quantization methods further confirms the efficacy of the proposed scheme.  
\end{abstract}

\begin{IEEEkeywords}
Federated Learning, Distributed Learning, Optimally Clipped Tensors And Vectors (OCTAV), Deterministic Quantization, Stochastic Quantization, Quantization Aware Training (QAT).
\end{IEEEkeywords}

\section{Introduction}

\begin{figure}[ht]
    \centering
    \includegraphics[width=1\linewidth]{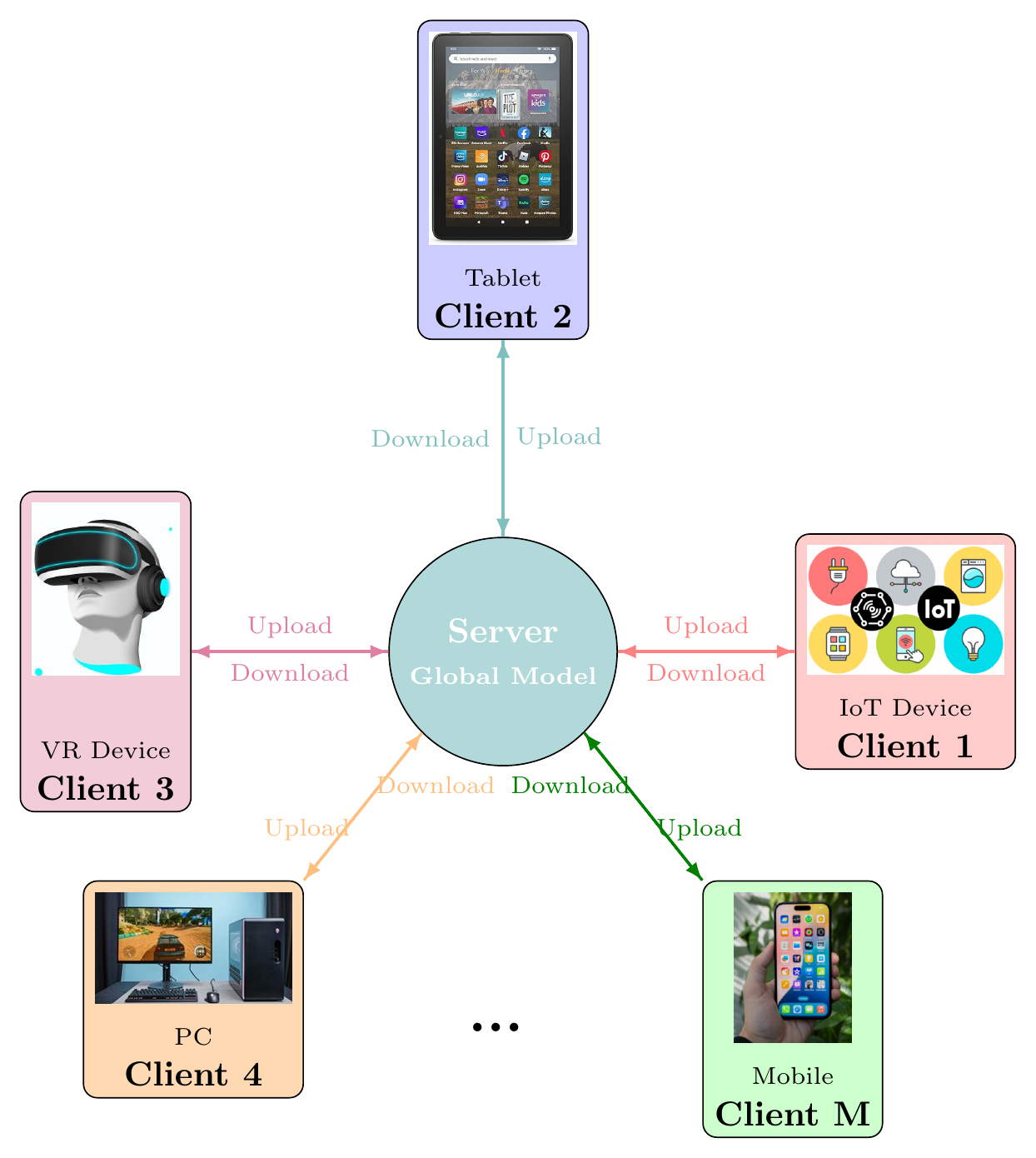}
    \caption{An illustration of a typical federated learning system.}
    \label{fig:FL}
\end{figure}

There has been extensive research on federated learning (FL) in which multiple decentralized devices/clients collaborate to develop a shared predictive model without transferring the raw data. This enhances data privacy while reducing the need for centralized data storage \cite{mcmahan2017communication}.  In a typical FL framework like Figure \ref{fig:FL}, the central server initializes a global model and shares it with a group of participating clients, such as smartphones, IoT devices, or organizational data centers. Each client performs local training on its private dataset, refining the global model using its data through optimization techniques like stochastic gradient descent (SGD). Once training is complete, clients send only the model updates—such as gradients or weight parameters—to the central server, avoiding the transmission of raw data. The server aggregates these updates using algorithms like Federated Averaging to create an improved global model, which is redistributed to clients for subsequent training rounds. This iterative process continues until the model converges to the desired level of accuracy. By keeping data localized and transmitting only essential updates, FL not only reduces communication overhead but also ensures privacy and scalability in distributed environments \cite{li2021survey}. The intentional asymmetry in the Figure \ref{fig:FL} of clients and the overall structure highlights the inherent heterogeneity characteristic of federated learning environments.

Federated learning is applied in areas like mobile keyboards (e.g., Google’s Gboard), voice recognition (e.g., Apple’s “Hey Siri”), healthcare (for secure analysis of medical records and images), and finance (for fraud detection and risk prediction), enabling collaborative model improvements while preserving data privacy \cite{kairouz2021advances}. 

Federated learning faces significant challenges related to communication bottlenecks and scalability due to the high volume of data exchanged between numerous devices and a central server \cite{reisizadeh2020fedpaq}. To combat these challenges, this paper introduces quantization techniques to alleviate communication load, enabling more efficient and scalable model training across distributed devices.

Compression techniques in neural networks are essential for reducing model size and computational costs, enabling deployment on resource-constrained devices while maintaining accuracy and efficiency, thus expanding the accessibility and scalability of deep learning applications \cite{han2015deep, jacob2018quantization,gholami2022survey}. Most existing literature primarily emphasizes memory savings and computational efficiency, yet communication constraints often pose a more significant challenge in distributed learning environments.
Different compression methods, including quantization, play significant roles in distributed learning by reducing the communication and/or computation overhead, which is crucial for efficiency in federated learning environments \cite{cao2023communication}. \\

The remainder of this paper is organized as follows: 
Section~\ref{Sec:ProposedAlgorithm} introduces the proposed federated learning algorithm, detailing the role of clipped uniform quantization and its integration into the training framework. 
Sections~\ref{SC: Clipped Uniform Quantizers} and \ref{SC: AggregationStrategies} delve into the mathematical formulation of the quantization process and aggregation strategies, followed by extensive simulation results in Section~\ref{Sec:Simulations} and concluding remarks in Section~\ref{Sec:Conclusion}.

\section{Proposed Federated Learning Algorithm}
\label{Sec:ProposedAlgorithm}
Algorithm \ref{Alg1} Shows the proposed federated learning setting. First, all clients start from the same initial model, then, after doing quantization aware training which uses optimal clipping threshold, they send the trained weights and also scaling factors to the server \footnote{The clipped quantized will be discussed in Section ~\ref{SC: Clipped Uniform Quantizers}. In case of aggregation by average square error, each client also sends the sum of square error of quantization errors of all the weights of each layer.}. After aggregation in the server by weighting based on proportion of data of each client (FedAvg) or averaging based on inverse of  mean of squared errors of quantization, the server sends back the full precision model. This process continues for $T$ rounds of training.  
Consider the federated averaging framework with $M$ clients, each holding a private dataset that is not shared. The collective goal is to minimize a global objective function under the constraints of data privacy. This objective is formalized as follows:

\begin{equation}
\label{eq:FedAvg}
\min_{z \in \mathbb{R}^d} f(z) = \min_{z \in \mathbb{R}^d} \sum_{i=1}^M \frac{n_i}{N} f_i(z),
\end{equation}

where $f_i(z) = \mathbb{E}_{L_i} [L_i(z, \xi_i)]$ denotes the differentiable loss function of client $i$, evaluated at data sample $\xi_i$ and model parameters $z$ in $d$ dimension. Here, $n_i$ represents the size of the dataset of client $i$, with the total size of all clients' datasets being $\sum_{i=1}^M n_i = N$. We propose two types of aggregation weighting in step 11 of Algorithm \ref{Alg1}. The first method follows the approach in \eqref{eq:FedAvg}, where client weights are aggregated based on their proportional contribution, as in Federated Averaging (FedAvg) \cite{mcmahan2017communication}. The second method utilizes the inverse of the average squared quantization error of the weights, offering an alternative aggregation strategy that will be detailed further in the subsequent sections.

\begin{algorithm}
\caption{The Proposed Federated Learning Algorithm}
\label{Alg1}
\begin{algorithmic}[1]
\State Every client downloads a neural network with the same structure on their local machine. Additionally, in case of FedAvg, each client reports the size of their local dataset to the server.
\For{$t = 1$ to $T$}
    \For{all the clients which are selected by the server for this round of training}
        \For{$\tau $ training steps/epochs}
           \State Find the optimal clipping scalar (or maximum of a tensor when we use maximum scalar quantization). For each layer, we have one scaling factor, and in case of FL based on average squared quantization error, one  error 
           for each layer.
           \State Perform clipping of the weights based on this scaling factors
           \State Perform quantization and dequantization based on number of bits and scaling factor 
            
        \EndFor

        \State Sends the K-bit weights and scaling factors to the server
    \EndFor
    \State Server obtains the aggregated/global model averaging based on FedAvg or averaging based on average squared quantization error
    \State Server sends the aggregated model to the clients
\EndFor
\end{algorithmic}
\end{algorithm}

We specifically focused on the quantization of weights during the uplink transmission (from clients to the server) for two primary reasons. First, the communication cost during the download phase is relatively low, as global parameters can be efficiently distributed to all clients via broadcasting. Second, quantizing weights on the server and sending them to clients results in all clients starting the next training round from identical weights. This uniformity negates the advantages of stochastic quantization, where diverse/stochastic weight realizations among clients can enhance model robustness.  
\section{Clipped Uniform Quantizers}
\label{SC: Clipped Uniform Quantizers}

\subsection{Clipping and two schemes of quantization}
While this paper focuses on the quantization of neural networks, it is worth noting that the theory of quantization in signal processing dates back to 1948 \cite{gray1998quantization}.This survey offers a comprehensive overview of the fundamental principles of quantization and highlights many of the widely used and promising techniques in the field. In low-precision quantization, converting floating-point values to integers inevitably introduces quantization noise, which is closely related to the dynamic range—the span between the maximum and minimum values that need to be represented. For a fixed \( b \)-bit integer format, narrowing this dynamic range can reduce the spacing between quantization levels, thereby minimizing quantization noise by enhancing precision. One effective way to limit the dynamic range is through clipping, which discards extreme values outside a certain threshold. However, this clipping process adds its own error by excluding valuable information from these outliers. Thus, finding an optimal clipping threshold involves balancing between reduced quantization noise and increased clipping error to achieve minimal information loss.

In this work, we explore the role of clipping in reducing quantization noise, aiming to enhance the precision of low-bit quantization schemes. As a preliminary step, we examined the statistical properties of tensor values in neural networks. Our measurements revealed that both weight and activation distributions often exhibit a bell-shaped pattern, as also observed in prior studies \cite{nagel2019data, banner2019post}. This observation implies that extreme values are infrequent compared to small ones, suggesting that clipping may discard only a minimal portion of information, while allowing more frequent values to benefit from higher precision.

In many cases, converting weights and activations to an 8-bit fixed-point format has minimal impact on model accuracy, as demonstrated in previous studies. However, reducing precision beyond this level often leads to a sharp decline in performance, emphasizing the need for an effective clipping strategy that minimizes information loss during quantization \cite{nagel2019data, banner2019post}.

 In \cite{migacz20178} Kullback-Leibler Divergence Measure (KLD) was proposed for clipping. The Kullback-Leibler Divergence (KLD) based method is a computationally intensive process, as it involves repeatedly assessing the KLD metric across a large set of possible clipping values. The algorithm identifies the optimal clipping threshold by selecting the value that yields the minimum divergence. Recent studies, such as those by Choi et al. \cite{choi2018pact} and Jung et al. \cite{jung2018joint}, introduced an approach to optimize activation clipping during training, while Wu et al. \cite{wu2018training} proposed a heuristic method to adjust clipping values based on target precision. In contrast, we use an analytical method to determine the optimal clipping threshold by minimizing the Mean Squared Error (MSE) based on the tensor's statistical distribution \cite{banner2018aciq, sakr2022optimal}.

For a continuous scalar \( x \), a clipped uniform quantizer first clips values outside the specified clipping range, then quantizes the remaining values into \( L \) levels, where \( L = 2^b \) for a \( b \)-bit quantizer.
 Without loss of generality, we assume the clipping thresholds, i.e., ${s}$ and ${-s}, s>0$ are symmetric\footnote{The asymmetric threshold can be easily achieved by introducing a bias term prior to quantization.}. Indeed, we aim to minimize the MSE between the original value and its quantized representation. Let \( X \) represents a random variable corresponding to the values being quantized (e.g., weights or activations), \( Q(X) \) denotes the quantized version of \( X \), and \( f(x) \) is the probability density function (PDF) describing the distribution of \( X \). The error is formally defined as:

\begin{align}
E\left[(X - Q(X))^2\right] &= 
\int_{-\infty}^{-s} f(x) \cdot (x + s)^2 \, dx \nonumber \\
&\quad + \sum_{i=0}^{2^b-1} \int_{-s + (i+1) \cdot \Delta}^{-s + i \cdot \Delta} f(x) \cdot (x - q_i)^2 \, dx \nonumber \\
&\quad + \int_{s}^{\infty} f(x) \cdot (x - s)^2 \, dx,
\end{align}
where \( q_i \), for \( i \in [0, 2^b-1] \), represents the quantization levels, typically chosen as the midpoints of the quantization intervals and the quantization step \(\Delta\), representing the interval between two consecutive quantized values, is defined as \(\Delta = \frac{2s}{2^b}\). The second term accounts for discretization, while the first and third terms correspond to the clipping error at the boundaries. An optimum threshold, $s$,  balances the trade-off between discretization and clipping noise.

The quantization process can be described as follows: 

\begin{enumerate}
    \item \textbf{Clipping:} First, the value \( x \) is clipped to the threshold, yielding \( \tilde{x} = \min(\max(x, -s), s) \).
 Hence, the question is how to find the threshold with a method which does not have high computational complexity?\\
    Banner et. al. \cite{banner2018aciq} derived a numerical equation/solution for optimal clipping values under both Gaussian and Laplacian noise distributions to minimize quantization noise. Al-Dhahir et. al. \cite{al1996uniform} investigated bit precision requirements for analog-to-digital converters for Gaussian signals, setting the quantizer's clipping level based on the input probability density function to achieve a specified distortion level. We use an online recursive method to determine the value of $s$ is \cite{sakr2022optimal}\\
    \begin{equation}
s_{n+1} = \frac{\sum_{x \in \Omega} \left[ |x| \cdot \mathbf{1}_{\{ |x| > s_n \}} \right]}{\frac{4 ^{- b}}{3} \sum_{x \in \Omega} \left[ \mathbf{1}_{\{ 0 < |x| \leq s_n \}} \right] + \sum_{x \in \Omega} \left[ \mathbf{1}_{\{ |x| > s_n \}} \right]}
\end{equation}

    which starts from an initial random guess \( s_1 \) and counts over a tensor or vector \( \Omega \) and then iteratively updating \( \{s_n\}_{n>1} \) until convergence. In our work, the algorithm converges within a maximum of 10 iterations.
    \item \textbf{Uniform quantization with $L$ levels}.\\ \textbf{Deterministic quantization} refers to a method where the input signal (or model weights, in the case of neural networks) is mapped to a fixed, predefined set of discrete values using a deterministic function. This means that for a given input, the quantized output will always be the same, and the mapping process does not involve randomness. In deterministic quantization, the error between the original value and its quantized representation is typically minimized using fixed thresholds or quantization levels.

\textbf{Stochastic quantization}, on the other hand, introduces randomness into the quantization process. Rather than directly assigning a fixed quantized value to the input, stochastic quantization adds noise or a probabilistic decision-making process. For example, the quantization levels might be chosen based on the probability distribution of the input value. This approach allows for a more flexible representation, potentially improving the robustness of the quantization process and reducing errors in certain scenarios, such as during the training of neural networks. For a uniform stochastic quantization\footnote{It's also called uniform dithered quantization.}, a uniform noise ranging between $(-\Delta /2,\Delta /2)$ is added to $x$ prior to quantization. Averaging over realizations of the stochastic quantization process effectively eliminates the bias inherent in deterministic quantization. An additional key advantage of stochastic quantization in federated learning is its impact on the starting point of client training. When the server sends the aggregated model to all clients, the weights are quantized, introducing randomness to their initialization. As a result, clients do not start training from identical weights, but rather from slightly varied initializations. This diversity in starting points, especially with a large number of clients, promotes a more robust and generalized model training process. As a result, the approach is less myopic and provides a broader, more comprehensive perspective. A thoughtful analogy provided by the second author compares this idea to a flock of birds navigating together. Instead of all birds flying in exactly the same direction, it may be beneficial for each bird to introduce some randomness in its path. This diversity allows the overall trend—essentially the average of all individual directions—to guide the flock more effectively toward an optimal course.  
\end{enumerate}

\subsection{Example: Stochastic vs Deterministic Quantization}
This example highlights the fundamental difference between deterministic and stochastic quantization, demonstrating how the latter reduces bias by probabilistically distributing values across quantization levels. While illustrated with a scalar value, the principle generalizes to the quantization of neural network weights, gradients, and activations, where stochastic methods enhance robustness and reduce cumulative quantization artifacts.\\
Consider quantizing \( x = 0.8 \) using a 2-bit uniform quantizer with levels \( q_0 = -1 \), \( q_1 = -0.33 \), \( q_2 = 0.33 \), and \( q_3 = 1 \) (\( \Delta = 0.66 \)).

\textbf{Deterministic Quantization}: \( x = 0.8 \) is mapped to the nearest level \( q_3 = 1 \), resulting in a quantization error:
\begin{equation}
e_{\text{det}} = x - Q(x) = 0.8 - 1 = -0.2
\end{equation}

\textbf{Stochastic Quantization}: as one possible ways of stochastic quantization, suppose \( x \) is probabilistically mapped to \( q_2 = 0.33 \) or \( q_3 = 1 \), with probabilities proportional to their distances:
\begin{equation}
P(q_2) = \frac{1 - 0.8}{\Delta} \approx 0.30, \quad P(q_3) = \frac{0.8 - 0.33}{\Delta} \approx 0.70
\end{equation}
The expected quantized value is:
\begin{equation}
\mathbb{E}[Q(x)] = (0.30 \cdot 0.33) + (0.70 \cdot 1) = 0.799
\end{equation}
The expected error becomes:
\begin{equation}
e_{\text{stoch}} = x - \mathbb{E}[Q(x)] = 0.8 - 0.799 = 0.001
\end{equation}

Stochastic quantization reduces bias by leveraging probabilities, achieving significantly lower error (\( e_{\text{stoch}} = 0.001 \)) compared to deterministic quantization (\( e_{\text{det}} = -0.2 \)). This makes it particularly effective in neural network training and federated learning, where cumulative quantization bias can degrade model performance.

\subsection{Averaging based on average squared quantization errors of weights}
 As it's not communication efficient to send the quantization error of every individual weights, we have found that sending average squared quantization errors of a tensor/vector (which might include several layers) will lead to the similar result as FedAavg. Let \( w_{ij,p} \) represent the de-quantized weights of layer \( i \) with P weights for client \( j \)  where de-quantized values are obtained by multiplying scale factor for that layer by the quantized weight of that neuron. Moreover, let \( e_{ij} \) represent the average squared quantization errors for that layer and client, i.e., \( e_{ij} = \frac{e_{ij,1} + e_{ij,2} + \dots + e_{ij,P}}{P} \).
The aggregated weight for neuron p of layer \( i \) of the server, denoted as \( \bar{w}_{i,p} \), is computed as:
\begin{equation}
\label{eq:eqAVGER}
\bar{w}_{i,p} = \frac{\sum_{j=1}^M\frac{w_{ij,p}}{e_{ij}}}{\sum_{j=1}^M \frac{1}{e_{ij}}}
\end{equation}

where:
\begin{itemize}
    \item \(\bar{w}_{i,p} \in \boldsymbol{\bar{w}}_{i,P} =[\bar{w}_{i,1},\bar{w}_{i,2},...,\bar{w}_{i,P}]\), \({w}_{ij,p} \in \boldsymbol{{w}}_{ij} =[{w}_{ij,1},{w}_{ij,2},...,{w}_{ij,P}] \)
    \item \( M \) is the total number of clients,
    \item \( \frac{1}{e_{ij}} \) is the weight assigned to the contribution of client \( j \) for layer \( i \), emphasizing smaller quantization errors,
    \item The denominator \( \sum_{j=1}^M \frac{1}{e_{ij}} \) normalizes the weights to ensure proper aggregation.
\end{itemize}

The same procedure is applied to all layers of the clients’ models on the server. Once aggregation is complete, the aggregated weights are sent back to all clients for the next round of training, and this process is repeated iteratively. In essence, this averaging approach assigns greater importance (or weight) to parameters with lower quantization errors, while those with higher quantization errors are given less emphasis. Additionally, one might consider employing Expectation-Maximization (EM) \cite{EM} to refine the clipped weights, replacing them with EM-derived values or a hybrid approach that combines Mean Squared Error (MSE) and EM for enhanced accuracy.

\section{Simulations Results}
\label{Sec:Simulations}

We utilize a convolutional neural network (CNN) architecture \footnote{The code to replicate all experiments is available at \url{https://github.com/zavareh1/ClippedQuantFL}} similar to the one described in \cite{mcmahan2017communication}. Table \ref{NN} shows the configuration of the model. The baseline is full-precision FedAvg with floating point weight transmission. In all the presented experiments, to demonstrate the effectiveness of the proposed methods, we have not fine-tuned them to achieve the highest possible accuracy, as our focus is on showcasing their viability rather than on precise accuracy comparisons. Therefore, we compared these configurations with the same hyper-parameters.

\begin{table}[h]
\centering
\caption{Configuration of the neural network employed for the MNIST dataset.}
\label{NN}
\begin{tabular}{|l|c|c|c|}
\hline
\textbf{Layer Type}       & \textbf{Kernel Size} & \textbf{Input Size}    & \textbf{Output Size}   \\ \hline
Conv2d (conv1)            & 3x3                  & 1 x 28 x 28            & 16 x 28 x 28           \\ \hline
BatchNorm2d (bn1)         & N/A                  & 16 x 28 x 28           & 16 x 28 x 28           \\ \hline
MaxPool2d (maxpool1)      & 2x2                  & 16 x 28 x 28           & 16 x 14 x 14           \\ \hline
Conv2d (conv2)            & 3x3                  & 16 x 14 x 14           & 16 x 14 x 14           \\ \hline
BatchNorm2d (bn2)         & N/A                  & 16 x 14 x 14           & 16 x 14 x 14           \\ \hline
MaxPool2d (maxpool2)      & 2x2                  & 16 x 14 x 14           & 16 x 7 x 7             \\ \hline
Linear (fc1)              & N/A                  & 784 (flattened)        & 100                    \\ \hline
BatchNorm1d (bn3)         & N/A                  & 100                    & 100                    \\ \hline
Linear (fc2)              & N/A                  & 100                    & 10                     \\ \hline
BatchNorm1d (bn4)         & N/A                  & 10                     & 10                     \\ \hline
Softmax                & N/A                  & 10                     & 10                     \\ \hline
\end{tabular}
\end{table}

In our simulations, we evaluated various communication schemes for the uplink transmission. Specifically, we explored an approach where the average squared quantization errors for each layer are computed and transmitted alongside the quantized weights to the server. This involves sending four values—corresponding to the average squared errors of each layer—encoded in 32 bits each. These error values are used to weight the contributions of each client's layers or neurons during server-side aggregation, based on the inverse of the average squared error for a layer divided by the sum of inverse errors across corresponding layers from other clients, i.e., equation (\ref{eq:eqAVGER}). Our results indicate that this error-weighted aggregation performs comparably to the conventional method of FedAvg.

Table \ref{ComSaving} quantifies the communication savings achieved for a single client in this setup. The column labeled \textbf{Model Weights} specifies the bit-width allocation for quantized weights. For example, "4-2-2-4" indicates that the weights in the first and last layers are quantized to 4 bits, while the middle layers use 2 bits. This allocation is informed by the observation that the first and last layers typically exhibit greater dynamic range, as evidenced by the histograms of their weights. This property aligns with common practices in the literature, which often recommend using higher precision (e.g., at least 8 bits) for the first and last  layers to preserve model accuracy.

\begin{table}[h]
\centering
\caption{Communication and memory savings for various weight quantization bit-width configurations}
\label{ComSaving}
\resizebox{\columnwidth}{!}{%
\begin{tabular}{|c|c|c|c|}
\hline
\textbf{Model Weights} & \textbf{FL with Clipping and Quantization of Weights} & \textbf{FL with Full Precision Weights} & \textbf{Communication Saving Times} \\
\hline
4-4-4-4 & $80848 \times 4 + 4 \times 32$ & $80848 \times 32$ & $\approx 8$ \\
\hline
4-2-2-4 & $144 \times 4 + 2304 \times 4 + 78400 \times 2 + 1000 \times 4 + 4 \times 32$ & $80848 \times 32$ & 15.53 \\
\hline
2-2-2-2 & $144 \times 2 + 2304 \times 2 + 78400 \times 2 + 1000 \times 2 + 4 \times 32$ & $80848 \times 32$ & 15.98 \\
\hline
2-1-1-2 & $144 \times 2 + 2304 \times 1 + 78400 \times 1 + 1000 \times 2 + 4 \times 32$ & $80848 \times 32$ & 31.12 \\
\hline
\end{tabular}%
}
\end{table}

\begin{figure}[ht]
    \centering
    \includegraphics[width=1\linewidth]{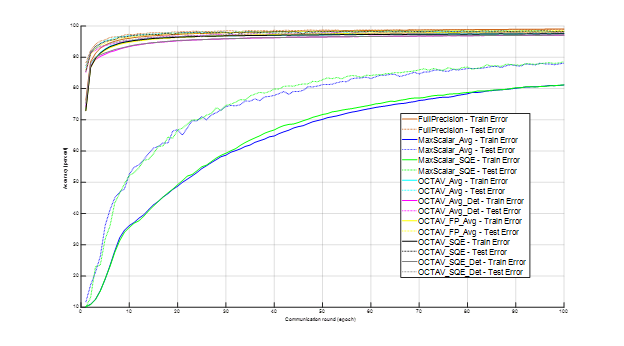}
    \caption{Communication rounds (Epoch) versus training and test accuracy (percent) for MNIST dataset.}
    \label{fig:fig2}
\end{figure}

 First, we conduct image classification on MNIST dataset, where 50000 samples have been used for training and 10000 samples for test of the model. For the MNIST dataset, to make the assumption of i.i.d, each client receives an equal number of images from every category, ensuring a uniform distribution across all classes. Each client out of 30 clients maintains its local dataset and trains a shared model independently, using stochastic gradient descent (SGD) as the optimizer with an initial learning rate of 0.01, a momentum of 0.9, and a weight decay of 1e-4. The local batch size is set to 64. These hyper-parameters are fixed for the next experiments. Figure \ref{fig:fig2} illustrates the performance of various methods under the "4-2-2-4" configuration. The Optimally Clipped Tensors And Vectors (OCTAV) framework includes several versions: averaging similar to FedAvg (OCTAV\_Avg), averaging based on the inverse of the average squared error of stochastic quantization (OCTAV\_SQE), and averaging based on the inverse of the average squared error of deterministic quantization (OCTAV\_SQE\_Det and OCTAV\_Avg\_Det), along with the full-precision version. Similar versions also exist for the maximum scalar approach. Except for the versions denoted with \_Det, all utilize stochastic quantization. The performance, evaluated on both MNIST and CIFAR-10 datasets, represent the average of 10 independent trials to ensure robustness and reproducibility.

From the figure, we observe that both the averaging method based on the number of clients and the method based on the inverse of the average squared error yield comparable performance. Additionally, methods employing stochastic quantization demonstrate superior robustness compared to their deterministic counterparts. This enhanced robustness highlights the benefits of incorporating stochasticity in the quantization process. Furthermore, approaches using optimal clipping significantly outperform those based on the maximum scalar, achieving performance levels close to that of full-precision communication in the uplink transmission.

\begin{figure}[ht]
    \centering
    \includegraphics[width=1\linewidth]{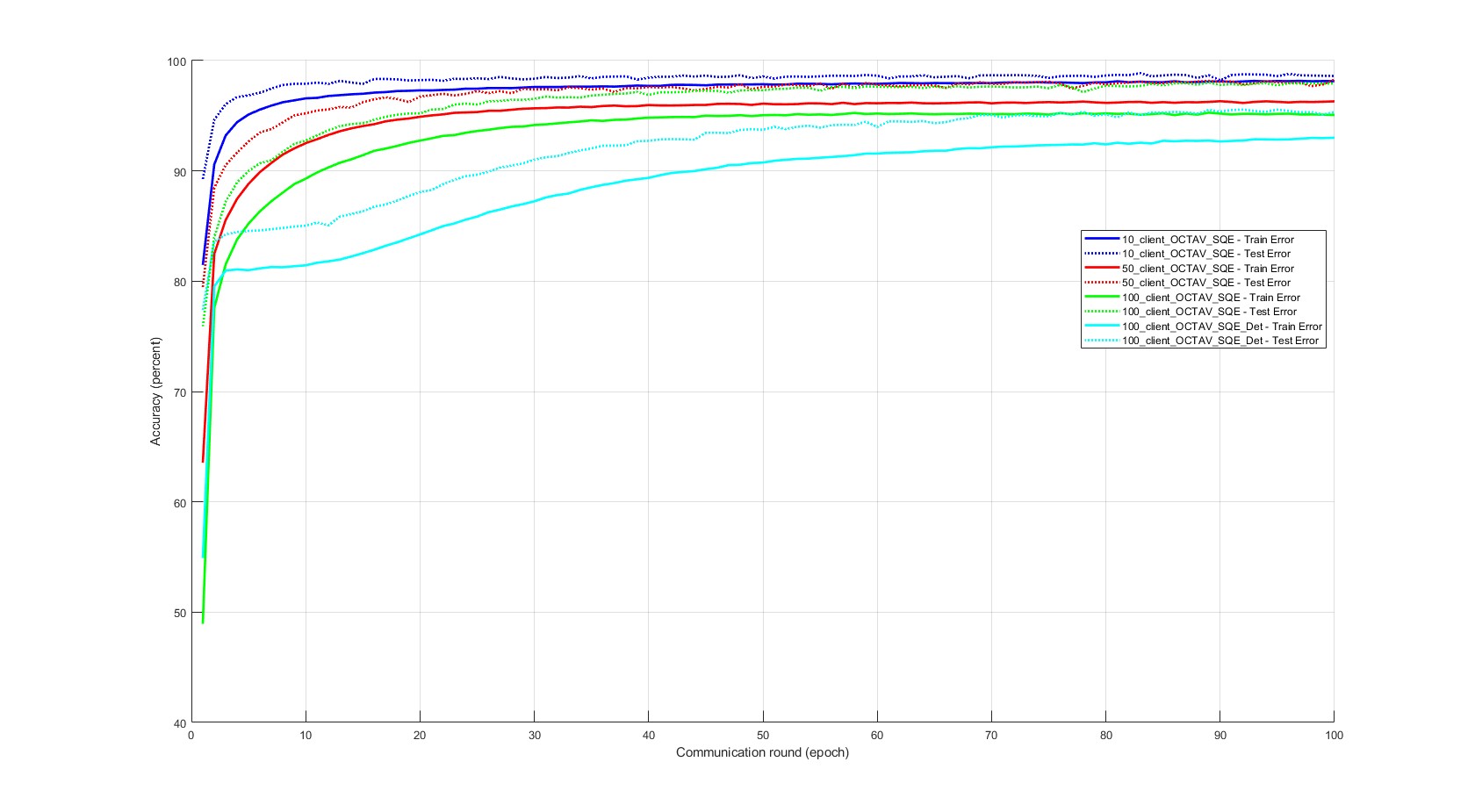}
    \caption{Communication rounds (Epoch) versus training and test accuracy (percent) for MNIST dataset for different number of clients.}
    \label{fig:fig3}
\end{figure}

In Figure \ref{fig:fig3}, the training and test errors are presented for the "2-2-2-2" configuration. The number of clients was varied across 10, 50, and 100. The results indicate that stochastic quantization significantly outperforms deterministic quantization when the number of clients is 100. As the number of clients increases, more communication rounds are required to achieve the same level of accuracy. Since averaging based on FedAvg produces results comparable to averaging based on the inverse of the average squared quantization error, and to avoid cluttering the figure, only the results for averaging based on the average squared quantization error are shown in this figure.

\begin{table}[h]
\centering
\caption{Configuration of the neural network employed for CIFAR-10 Dataset}
\label{NN_cifar}
\begin{tabular}{|l|c|c|c|}
\hline
\textbf{Layer Type}       & \textbf{Kernel Size} & \textbf{Input Size}    & \textbf{Output Size}   \\ \hline
Conv2d (conv1)            & 3x3                  & 3 x 32 x 32            & 16 x 32 x 32           \\ \hline
BatchNorm2d (bn1)         & N/A                  & 16 x 32 x 32           & 16 x 32 x 32           \\ \hline
MaxPool2d (maxpool1)      & 2x2                  & 16 x 32 x 32           & 16 x 16 x 16           \\ \hline
Conv2d (conv2)            & 3x3                  & 16 x 16 x 16           & 16 x 16 x 16           \\ \hline
BatchNorm2d (bn2)         & N/A                  & 16 x 16 x 16           & 16 x 16 x 16           \\ \hline
MaxPool2d (maxpool2)      & 2x2                  & 16 x 16 x 16           & 16 x 8 x 8             \\ \hline
Linear (fc1)              & N/A                  & 1024 (flattened)       & 100                    \\ \hline
BatchNorm1d (bn3)         & N/A                  & 100                    & 100                    \\ \hline
Linear (fc2)              & N/A                  & 100                    & 10                     \\ \hline
BatchNorm1d (bn4)         & N/A                  & 10                     & 10                     \\ \hline
Softmax                & N/A                  & 10                     & 10                     \\ \hline
\end{tabular}
\end{table}

\begin{figure}[ht]
    \centering
    \includegraphics[width=1\linewidth]{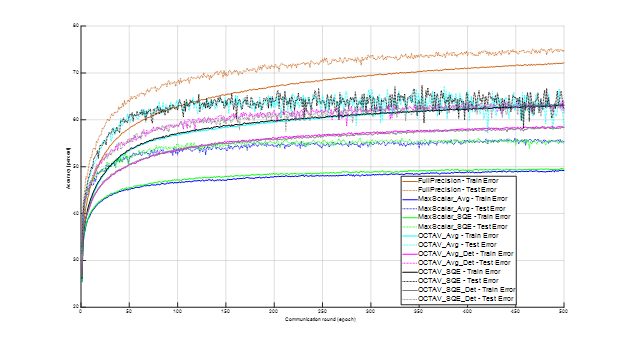}
    \caption{Communication rounds (Epoch) versus training and test accuracy (percent) for CIFAR10 dataset.}
    \label{fig:fig4}
\end{figure}

\begin{figure}[ht]
    \centering
    \includegraphics[width=1\linewidth]{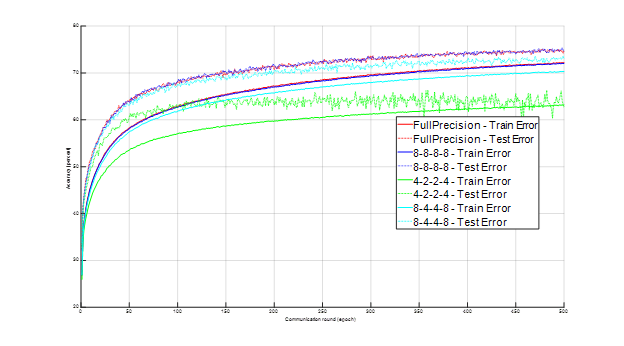}
    \caption{Comparison of training and test accuracy (percentage) across communication rounds (epochs) for different neural network configurations on the CIFAR-10 dataset.}
    \label{fig:fig5}
\end{figure}

In the next federated learning framework, i.e. Figure \ref{fig:fig4}, we focus on training a neural network of Table \ref{NN_cifar} collaboratively across 20 clients with configuration "4-2-2-4" using the CIFAR-10 dataset, a benchmark dataset consisting of 60,000 32x32 color images across 10 classes. The dataset is divided into 50,000 training samples and 10,000 test samples, which are distributed as independent and identically distributed (IID). The similar behavior like the MNIST one has been observed except while the accuracy of the quantized model on the MNIST dataset is comparable to that of the full-precision model, a substantial difference is observed in the case of CIFAR-10. Since CIFAR-10 is more challenging than MNIST, it may be preferable to use a configuration like "8-8-8-8" instead of "4-2-2-4" to maintain higher accuracy. As demonstrated in Figure \ref{fig:fig5}, the "8-8-8-8" configuration achieves performance on the CIFAR-10 dataset that is similar to the full-precision model. Moreover, with the "8-8-8-8" configuration, the communication and memory requirements are reduced to approximately one-fourth of those for the full-precision model, highlighting its efficiency in resource-constrained scenarios. The "8-4-4-8" configuration also achieves near full-precision accuracy, making it a viable alternative

\begin{figure}[ht]
    \centering
    \includegraphics[width=1\linewidth]{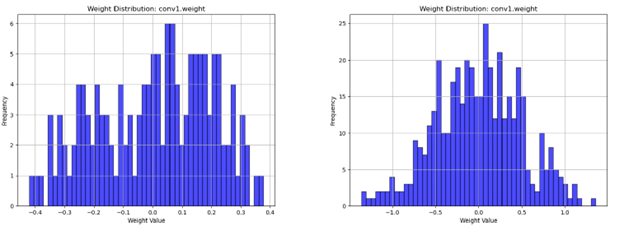}
    \caption{Comparison of histogram of weights of the first layers of configurations on the MNIST and CIFAR-10 datasets.}
    \label{fig:hist}
\end{figure}

\begin{figure}[ht]
    \centering
    \includegraphics[width=1\linewidth]{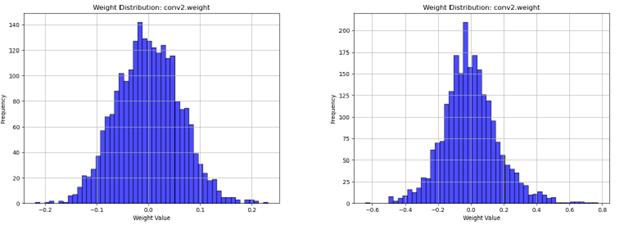}
    \caption{Comparison of histogram of weights of the second layers of configurations on MNIST and CIFAR-10 datasets.}
    \label{fig:hist2}
\end{figure}

\begin{figure}[ht]
    \centering
    \includegraphics[width=1\linewidth]{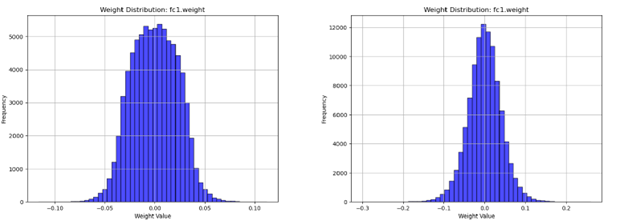}
    \caption{Comparison of histogram of weights of the third layers of configurations on the MNIST and CIFAR-10 datasets.}
    \label{fig:hist3}
\end{figure}

\begin{figure}[ht]
    \centering
    \includegraphics[width=1\linewidth]{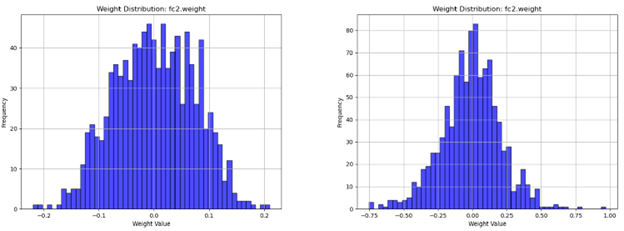}
    \caption{Comparison of histogram of weights of fourth layers of configurations on the MNIST and CIFAR-10 datasets.}
    \label{fig:hist4}
\end{figure}

It is worth noting that the dynamic ranges of the first and last layers in neural networks typically exhibit higher variability compared to intermediate layers. Consequently, it has been often recommended allocating at least 8-bit precision for these boundary layers to preserve critical information and maintain model performance. As an example, in Figures \ref{fig:hist} to \ref{fig:hist4} the histograms of the first layers to the fourth layers of the trained configurations on the server for MNIST and CIFAR-10 are shown after 100, and 500 rounds of training, respectively. The left one shows the histograms for the configuration on MNIST, and the right one shows the histograms for the configuration on CIFAR-10. We see that for CIFAR it has higher dynamic range.\\


Moreover, in layerwise quantization, the same clipping range is applied across all filters within the same layer. However, this approach can lead to poor quantization resolution for channels with narrow value distributions \footnote{For a discussion on various quantization granularities, we refer the reader to \cite{gholami2022survey}.}. By contrast, channelwise quantization assigns distinct clipping ranges to each channel, allowing for improved quantization resolution. Indeed, as an alternative, instead of considering one scale factor for each layer, which is per tensor, one may like to do per channel quantization which requires more computation to compute scale factor for each individual channel. Though this, in turn, would add a bit more communication, the significant load would be on computation. A combination of increasing bit-width of some layers and per channel quantization also works.  This frameworks highlights the efficacy of integrating quantization-aware training with federated optimization in resource-constrained settings.


%

\section{Conclusion}
\label{Sec:Conclusion}
In this paper, we introduced a novel approach to enhance communication efficiency in federated learning through clipped uniform quantization. By integrating optimal clipping thresholds and stochastic quantization schemes, our method significantly reduces the communication overhead while maintaining near-full-precision model accuracy. The proposed framework also offers a privacy-preserving benefit by mitigating the need to disclose client-specific dataset sizes to the central server.

This work underscores the potential of integrating advanced quantization techniques into federated learning to address key challenges such as scalability and resource constraints. For future work, while our framework focuses on quantization of weights, exploring the quantization of activations and gradients could provide a holistic approach to communication-efficient federated learning. Additionally, integrating adaptive bit-width quantization that dynamically adjusts based on the model's convergence state or dataset heterogeneity could further optimize resource utilization.

\bibliographystyle{IEEEtran}
\bibliography{references}

\end{document}